\documentclass{article}

\usepackage{microtype}
\usepackage{graphicx}
\usepackage{subfigure}
\usepackage{booktabs} 
    \usepackage{xcolor}
\usepackage{amsmath}
\usepackage[section]{placeins}

\usepackage{amssymb}

\usepackage{hyperref}

\usepackage[accepted]{icml2018}

\icmltitlerunning{Defending Against Adversarial Samples Using An Entire GAN
}

\begin{document}

\twocolumn[
\icmltitle{Defending Against Adversarial Attacks by Leveraging an Entire GAN}



\icmlsetsymbol{equal}{*}

\begin{icmlauthorlist}
\icmlauthor{Gokula Krishnan Santhanam}{eth}
\icmlauthor{Paulina Grnarova}{eth}
\end{icmlauthorlist}

\icmlaffiliation{eth}{Department of Computer Science, ETH Zurich, Zurich, Switzerland}

\icmlcorrespondingauthor{Gokula Krishnan Santhanam}{sgokula@ethz.ch}

\icmlkeywords{GAN, Adversarial Attacks, Classifiers}

\vskip 0.3in
]



\printAffiliationsAndNotice{}  

\begin{abstract}
Recent work has shown that state-of-the-art models are highly vulnerable to adversarial perturbations of the input. We propose \emph{cowboy}, an approach to detecting and defending against adversarial attacks by using both the discriminator and generator of a GAN trained on the same dataset. We show that the discriminator consistently scores the adversarial samples lower than the real samples across multiple attacks and datasets. We provide empirical evidence that adversarial samples lie outside of the data manifold learned by the GAN. Based on this, we  propose a cleaning method which uses both the discriminator and generator of the GAN to project the samples back onto the data manifold. This cleaning procedure is independent of the classifier and type of attack and thus can be deployed in existing systems. 
\end{abstract}

\section{Introduction}
\label{intro}


Recent work on adversarial attacks \cite{orig-adv, fgsm, basic-iterative-method} has shown that neural networks are brittle and can easily be confused by small imperceptible modifications of the input. While these modifications do not affect human perception, they cause misclassification of samples in deep-learning models creating a threat for security and safety. 

Current defenses against adversarial samples follow two lines of work: (1) Modifying the training procedure of the classifier to make it more robust by either modifying the training data (adversarial learning \cite{orig-adv, fgsm}) or blocking gradient pathways (e.g. defensive distillation \cite{papernot2016distillation}) and (2) removing the adversarial perturbations from the input \cite{meng2017magnet, defense-gan, pixel-defend}.

We present \emph{cowboy}, an approach to detecting and defending against adversarial samples that is agnostic to the target classifier and its training procedure and is independent of the attack method. 
We leverage the power of a Generative Adversarial Network (GAN) \cite{gan} trained only on the real data, and use both the discriminator and generator to successfully defend against various attacks. 

A crucial step towards tackling adversarial samples is the ability to distinguish them from real samples. We postulate that adversarial samples lie outside of the data manifold and that the discriminator can effectively detect this. We test this hypothesis by conducting experiments using different attacks (Fast Gradient Sign Method (FGSM) \cite{fgsm}, Basic Iterative Method (BIM) \cite{basic-iterative-method}, Virtual Adversary Method (VAM) \cite{virtual-adversary}, Projected Gradient Descent Method (PGDM) \cite{projected-gradient} and Momentum Iterative Method (MIM) \cite{mometum-iterative-attack}) on MNIST, Fashion-MNIST, CIFAR-10, and SVHN datasets and show that the discriminator consistently scores the adversarial samples lower than the original samples. 

Once detected, the adversarial samples can be cleaned by projecting them back to the data manifold learned by the GAN and pushing them to a high scoring region guided by the score of the discriminator. When the cleaning procedure is used as a pre-processing step on the samples before given to the classifier, the accuracy of classification improves from $0.02\%$ to $0.81\%$ for one of the worst attack. 

Our main contributions can be summarized as:
\begin{itemize}
\item We empirically show that adversarial samples lie outside of the data manifold learned by a GAN that has been trained on the same dataset. 
\item We propose a simple procedure that uses both the generator and the discriminator to successfully detect and clean adversarial samples.
\end{itemize}  

The key feature of our framework is that both the GAN and classifier are not modified nor are shown adversarial samples during training. This allows our approach to generalize to multiple attacks and datasets with only a few lines of code. This makes our framework an easy to plug-in step in already deployed classification pipelines

The rest of the paper is organized as follows. Necessary background including various defense mechanisms and attack models that we use are introduced in Section 2. In Section 3, we describe and motivate \emph{cowboy}, a simple, yet effective defense algorithm. Finally, Section 4 presents and discusses experimental results for both detection and cleaning under different attack methods for generating adversarial samples across 4 different datasets.

\section{Background and Related Work}
\paragraph{Attack Strategies.} Adversarial samples were first reported by \cite{orig-adv}. Since then many attack strategies have been proposed, which can broadly be classified into black-box and white-box attacks. White-box strategies have access to all the weights and gradients of the classifier. Black-box strategies on the other hand, have access only to the predictions of the network. Even though this might make the attacks more difficult, it enables successful attacks to be transferable to many classifiers. 

In this work, an attack strategy is an algorithm that perturbs an original sample $X \in \mathbf{R}^n$ to an adversarial sample $X_{adv} = X + \delta \in \mathbf{R}^n$ in a way that the change is undetectable to the human eye. The change is measured by the $\it{l}_{\infty}$ norm of the perturbation, denoted by $\epsilon$. We focus on the following  attack methods:
\subparagraph{Fast Gradient Sign Method \cite{fgsm} }
The Fast Gradient Sign Method (FGSM) generates adversarial images using 
\begin{equation*}\label{eq:fgsm_adv}
	X^{adv}=X + \epsilon \cdot sign(\bigtriangledown_{X}J(X,y_{true})),
\end{equation*}
 where $\epsilon$ controls the perturbation's amplitude, with smaller values creating imperceptible perturbations. This attack moves all the pixels of the input $X$ in the direction of the gradient simultaneously. This method is easy to implement and inexpensive to compute but is easily detected by current methods.

\subparagraph{Basic Iterative Method \cite{basic-iterative-method}} 
Basic Iterative Method (BIM) is a variant of the FGSM method. It applies FGSM  multiple times with a smaller step size. The adversarial examples are computed as 
\begin{equation*}
    X^{adv}_0 = X,  X^{adv}_{n+1} = X^{adv}_{n}+ \alpha \cdot sign(\bigtriangledown_{X}J(X^{adv}_{n+1},y_{true})).
\end{equation*}
In practice, $\alpha$ is smaller when compared to $\epsilon$ of FGSM. We also clip the updates such that the adversarial samples lie within the $\epsilon$-ball of $X$.

\subparagraph{Momentum Iterative Method \cite{mometum-iterative-attack}}
The Momentum Iterative Method (MIM) is a variant of BIM that exploits momentum \cite{momentum} when updating $X^{adv}_n$. This results in adversarial samples of superior quality.\footnote{This method won the first places in NIPS 2017 Non-targeted Adversarial Attack and Targeted Adversarial Attack competitions.}

\subparagraph{Projected Gradient Descent Method \cite{projected-gradient}}
Projected Gradient Descent Method(PGDM) is an optimization based attack which finds a point, $X^{adv}$, that lies within the $\epsilon$-ball of $X$ such that it minimizes the probability of the true label $y_{true}$.

\subparagraph{Virtual Adversary Method \cite{virtual-adversary}}
Virtual Adversary Method (VAM) uses Local Distributional Smoothness (LDS), defined as the negative sensitivity of the model distribution $p(y|x, \theta)$ with respect
to the perturbation of $X$, measured in terms of KL divergence. It exploits LDS to find the direction in which the model is most sensitive to perturbations to generate adversarial samples. 

\paragraph{Defense Strategies.} Common approaches to defending against adversarial attacks include training with adversarial samples \cite{fgsm}, modifying the classifier to make it more robust to attacks \cite{papernot2016distillation}, label-smoothing \cite{warde201611}, as well as using auxiliary networks to pre-process and clean the samples \cite{dae-cleaning-for-adv-samples, meng2017magnet}. In the context of purifying the samples, a new line of work focuses on using generative models to detect and defend against adversarial attacks. Models like RBMs and PixelRNNs \cite{pixel-rnn} have been used to varying degrees of success as well. 

Generative Adversarial Networks (GANs) \cite{gan} are an interesting prospect in defending against adversarial attacks. Even though training GANs is still notoriously difficult, there have been successful efforts in this direction. Defense-GAN \cite{defense-gan} uses the generator in order to sanitize inputs before passing them to the classifier. APE-GAN \cite{ape-gan} modifies a GAN such that the generator learns to clean the adversarial samples, whereas the discriminator tries to discriminate between real and adversarial input. While these approaches focus only on the generative part of the network, we use both the generator and the discriminator as means to detecting and protecting against attacks.
To our best knowledge, there has not been any prior work on using an unmodified discriminator for this task.

\section{Cowboy}
\label{sec:cowboy}

Our goal is twofold: we want to \textit{(i)} be able to identify whether an input is adversarial or not, and \textit{(ii)} clean it from its adversarial noise if it is. In order to do so, we aim at combining information drawn from two sources, respectively the generator $G$ and the discriminator $D$ of a GAN trained on the same dataset as our initial classifier. 

\subsection{Classifiers learn discriminative surfaces}
\label{sec:classifier}
\begin{figure}[h]
\vskip 0.2in
\begin{center}
\centerline{\includegraphics[width=\columnwidth]{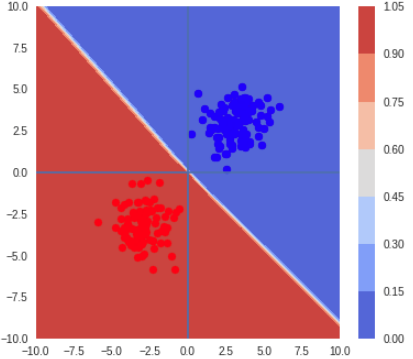}}
\caption{Linear discriminative surface learned by a classifier on a Mixture of two Gaussians.}
\label{fig:c-contour}
\end{center}
\vskip -0.2in
\end{figure}

As noted in \cite{fgsm}, common activation functions used in Deep Learning like Sigmoid, ReLU and Tanh are either piece-wise linear or used in regions where they have linear-like behavior. As a result, classifiers modeled as DNNs tend to learn hypersurfaces that split the output space into finitely many disjoint open subspaces. We can see this clearly in the case of a mixture of two Gaussians respectively centered at $(3, 3)$ and $(-3, -3)$ and with variance 1, as shown in Figure~\ref{fig:c-contour}. It is also interesting to note that the classifier assigns high probability to regions where it has never seen samples before, showing that the classifier has no notion of the data manifold. The classifier only learns a discriminative surface to maximize classification accuracy. 

\subsection{Adversarial Examples lie outside the Data Manifold}
\label{sec:outside}
There has been work \citep{adv-real-not-twins, pixel-defend, stat-detect-of-adv-samples} arguing that adversarial perturbations result in samples lying outside of the data manifold. Since the classifier has no knowledge of the manifold, this can result in misclassification with very high confidence of the classifier. We can consider this to be akin to testing out-of-distribution samples, which naturally results in poor test-time performance. This suggests the use of defense mechanisms involving the support of the real distribution $p_{data}$.

\subsection{$D$ can detect adversarial attacks}
\label{sec:D_is_great} 
What kind of information of $p_{data}$ is captured by the discriminator $D$? Consider the following two claims:
\begin{enumerate}
\item There exists $0<\delta\ll 1$ such that the support of $p_{data}$ is contained in $D^{-1}([1-\delta,1])$.
\item If $x$ in a real sample, then corrupting it into an adversarial sample $\tilde{x}$ using common adversarial attack methods would yield $D(\tilde{x})\ll 1$. 
\end{enumerate}

The first of these claims is equivalent to saying that testing if a sample $x$ is real by asking whether $1-\delta\leq D(x)\leq 1$ would not yield any false negative, which can also be formulated as an absence of mode-collapse for the discriminator $D$. 

This behaviour can be understood by reasoning about the dynamics of learning in the GAN's min-max game:
\begin{multline}
\label{eq:minimaxgame-definition}
\min_G \max_D V(D, G) = \mathbb{E}_{x \sim p_{\text{data}}(x)}[\log D(x)]+\\  
\mathbb{E}_{z \sim p_{z}(z)}[\log (1 - D(G(z)))].
\end{multline}
The discriminator has to assign higher scores to regions around the data points as these are shown as positive examples and doing otherwise would increase its loss.

We also assess the validity of this assumption empirically by drawing violin plots of the values of $D$ for a variety of datasets in Figure~\ref{fig:violins}, which appears to hold true for each dataset. This allows us to consider claim $1$ as being valid, under the assumption that the discriminator did not undergo mode collapse. 

\begin{figure}[h]
\vskip 0.2in
\begin{center}
\centerline{\includegraphics[width=\columnwidth]{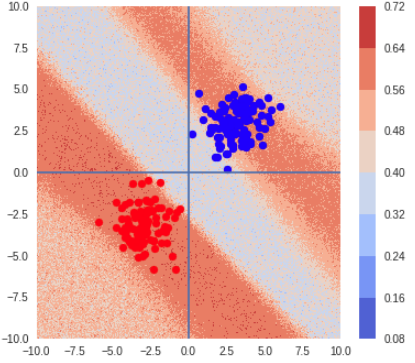}}
\caption{Score of the discriminator $D$ of a GAN trained on a mixture of two Gaussians.}
\label{fig:d-contour}
\end{center}
\vskip -0.2in
\end{figure}

However, what can be said about false positives for this test?, \textit{i.e.}, is there a consequent part of the space outside of the data manifold where $D$ scores high? As argued in Section~\ref{sec:classifier}, a classifier tends to only learn discriminative surfaces in order to maximize classification accuracy, and hence $D$ might assign high scores to the presence of real data in regions where it has never seen any sample, either real or from $G$. Indeed, this is what seems to happen in Figure~\ref{fig:d-contour}.

Therefore, in order to assess whether $D$ scoring low would give a good proxy for the detection of adversarial samples, it is both \textit{necessary} and \textit{sufficient} to assess the validity of claim $2$, i.e. that an adversarial sample is unlikely to live in the region of the space outside of the data manifold where $D$ scores high. Indeed, adversarial examples lie outside the data manifold (see Section~\ref{sec:outside}), and the data manifold is contained in the region where $D$ scores high.

Similarly as we did for claim $1$, we assess the validity of claim $2$ empirically by drawing violin plots of the values of $D$ on a variety of adversarial attacks. Figure~\ref{fig:violins} shows that adversarial samples get assigned a low score by $D$ 

This demonstrates that $D$ scoring low provides a good proxy for the detection of adversarial attacks.

\subsection{Estimating a good threshold}
The concepts discussed in the previous section are agnostic w.r.t. which attack is used. This means that if two completely different attacks generate adversarial samples in the same region, then scores of $D$ for these samples would be similar. This implies that we can obtain a good estimate for the threshold by using any attack method on the trained classifier. 
To do this, we can generate adversarial samples\footnote{Let's define adversarial samples as those that cause at least a 20\% (arbitrarily chosen) drop in accuracy} using an arbitrary attack method (say FGSM) for each (or a subset) data point in the dataset. A simple threshold could then be the average discriminator score across all adversarial samples. Note that although choosing the maximum instead of the average could be very badly affected by an outlier, one can also simply choose an $L^p$ average $((1/p)\sum(\cdot)^p)^{1/p}$ for some $p\in [1,\infty)$ as a way to interpolate between the average and the maximum.

\subsection{Taming Adversarial Samples}
\begin{center}
\begin{table*}[ht]
\caption{Classifier Accuracies (5,000 cleaning steps): The classifier accuracy on the adversarial samples increases significantly when the samples have been preprocessed with the Cowboy cleaning method.}
{\small
\hfill{}
\begin{tabular}{@{\extracolsep{1.5pt}}lrrrrrrrrrrrr}
\toprule
ATTACK & \multicolumn{3}{c}{ MNIST}  & \multicolumn{3}{c}{F-MNIST} & \multicolumn{3}{c}{CIFAR-10} & \multicolumn{3}{c}{SVHN} \\
\cline{2-4} \cline{5-7} \cline{8-10} \cline{11-13} \\
 & Original &  Adv &  Clean  & Original & Adv & Clean & Original & Adv & Clean & Original & Adv & Clean  \\
\midrule
FGSM ($l_{\infty})$ & 0.97 & 0.12 & 0.78        & 0.89 & 0.11 & 0.44        & 0.78 & 0.30 & 0.53        & 0.87 & 0.34 & 0.72 \\
PGDM ($l_{\infty})$ & 0.97 & 0.02 & 0.81        & 0.89 & 0.07 & 0.48        & 0.78 & 0.06 & 0.53        & 0.87 & 0.13 & 0.60 \\
BIM ($l_{\infty})$  & 0.97 & 0.02 & 0.15        & 0.89 & 0.07 & 0.13        & 0.78 & 0.20 & 0.47        & 0.87 & 0.16 & 0.49 \\
MIM ($l_{\infty})$  & 0.97 & 0.03 & 0.45        & 0.89 & 0.08 & 0.32        & 0.78 & 0.21 & 0.43        & 0.87 & 0.15 & 0.48 \\
VAM                 & 0.97 & 0.32 & 0.43        & 0.89 & 0.50 & 0.34        & 0.78 & 0.45 & 0.56        & 0.87 & 0.67 & 0.76 \\
\bottomrule
\end{tabular}}
\hfill{}
\label{tb:tablename}
\end{table*}
\end{center}

\begin{algorithm}[tb]
   \caption{Cleaning Adversarial Inputs}
   \label{cleaning-algo}
\begin{algorithmic}
   \STATE {\bfseries Input:} adversarial input $x$, learning rate $\eta$
   \STATE Sample $z_0$ from noise prior $p_z(z)$
   \FOR{$i=1$ {\bfseries to} $m$}
    \STATE Update $z_{i} \leftarrow z_{i-1} - \eta \nabla_{z}\mathcal{L}(z_{i-1})$
   \ENDFOR
   \STATE \textbf{return} $G(z_m)$, the cleaned image
\end{algorithmic}
\end{algorithm}
Once an input has been detected as adversarial, we can attempt to clean this input from its adversarial noise.  In order to do so, \citet{defense-gan} and similar work suggest to find an optimal $L^2$-reconstruction of $x$ by $G$:
\begin{equation}
 z^* = \arg\min_z \parallel G(z) - x \parallel_{2}^{2}.
\end{equation}

However, these approaches use only the generator $G$ of the GAN.

As already argued in the previous sections, the discriminator $D$ contains important information about the data manifold. How can we exploit this information in order to improve reconstruction? The probability that a reconstructed input $G(z)$ lies on the data manifold can be modelled by $D$ as $D(G(z))$, whose log-likelihood is then $\log(D(G(z)))$. Moreover, it is natural to interpret a re-scaled $L^2$-reconstruction term $\frac{1}{2\sigma^2}\Vert G(z)-x\Vert_2^2$ as the negative log-likelihood of an isotropic Gaussian of variance $\sigma^2$ centered at $x$, since the logarithm and the exponential cancel. A desired reconstruction term would aim at maximizing the likelihood that $G(z)$ be close to $x$, while lying on the data manifold. If we model these two events as being independent and having probability densities respectively $\mathcal{N}(x,\sigma^2 I_n)$ and $D(G(z))$ $-$ where $\sigma$ is a hyper-parameter of the model $-$, then this leads us to maximizing the following log-likelihood:
\begin{equation}
\log\left(\frac{1}{(2\pi\sigma^2)^{n/2}}e^{-\frac{1}{2\sigma^2}\Vert G(z)-x\Vert_2^2}\right)+\log(D(G(z))),
\end{equation}
which is strictly equivalent to minimizing the following quantity:
\begin{equation}
\mathcal{L}(z):=\frac{1}{2\sigma^2}\Vert G(z)-x\Vert_2^2-\log(D(G(z))).
\end{equation}
Note that $\sigma$ trades-off between exact reconstruction ($G(z)=x$) and realness of the reconstruction ($D(G(z))=1$). 

We can now find a minimizer $z^*$ of $\mathcal{L}$ by gradient-descent, as in \citet{defense-gan}, while keeping the generator and the discriminator fixed. We experiment with different values of $\sigma$ and discuss its effect in \ref{results}. The two terms composing our reconstruction loss $\mathcal{L}$ ensure that the cleaned image is similar to the input and lies on the data manifold, which is achieved without evaluating any likelihood over $p_{data}$ explicitly.




\section{Experiments}

\paragraph{Experimental Setup.} We conduct experiments on MNIST, Fashion-MNIST \cite{fashion-mnist}, CIFAR-10 and SVHN. We use the Fast Gradient Sign Method (FGSM), Basic Iterative Method (BIM), Virtual Adversary Method (VAM), Projected Gradient Descent Method (PGDM) and Momentum Iterative Method (MIM) to mount attacks on the classifier. 

The classifier is a simple CNN and the GAN is based on the DCGAN family of architectures \cite{dcgan}. Batch Normalization \cite{batch-norm} and Dropout \cite{dropout} are used  only for the classifier. The weights are initialized using the Xavier Initializer \cite{xav-init}. Other than normalizing the inputs to lie between -1 and 1, no special pre- or post-processing steps are performed. The models are implemented in TensorFlow \cite{2016arXiv160304467A} and the implementations of the adversarial attacks are based on CleverHans \cite{papernot2017cleverhans}\footnote{codebase will be released after the review process}. Both the classifier and the GAN for SVHN and CIFAR-10 are trained for 40K steps, whereas the ones for MNIST and Fashion-MNIST are trained for 2K steps. Once training is complete, we test on the entire respective test sets. Further details on the hyperparameters along with the specific architectures can be found in the codebase and supplementary material.

\begin{figure*}[h]
\centering
\begin{tabular}{cc}
  \includegraphics[width=75mm]{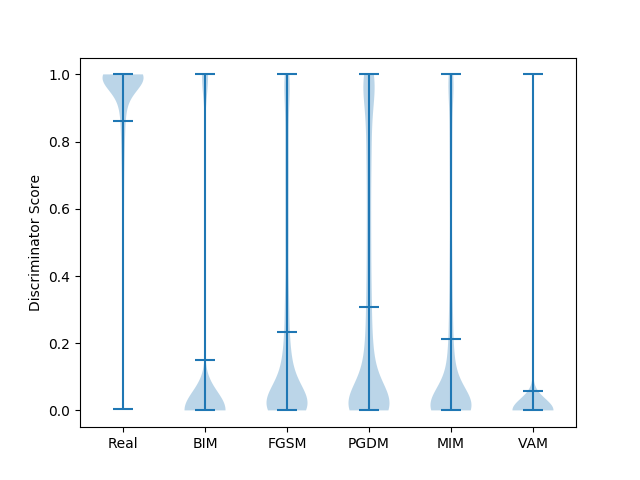} &   \includegraphics[width=75mm]{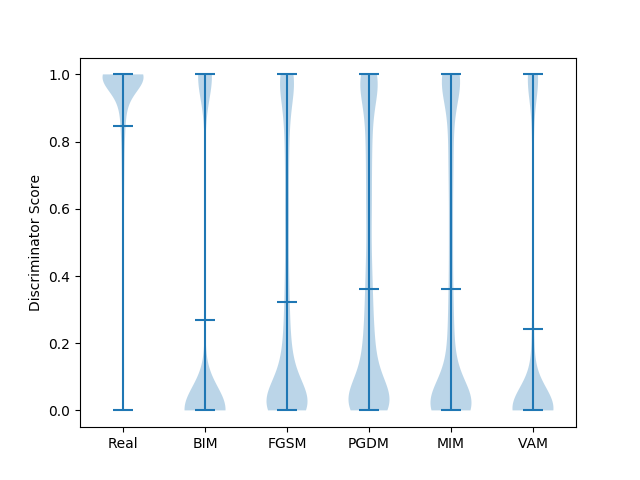} \\
(a) MNIST & (b) Fashion-MNIST \\[6pt]
 \includegraphics[width=75mm]{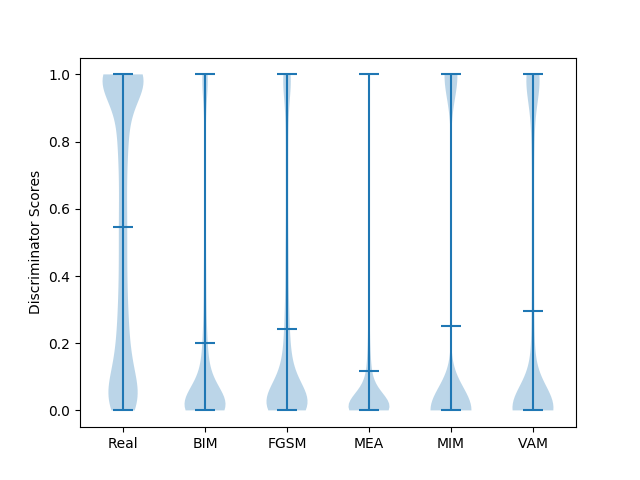} &   \includegraphics[width=75mm]{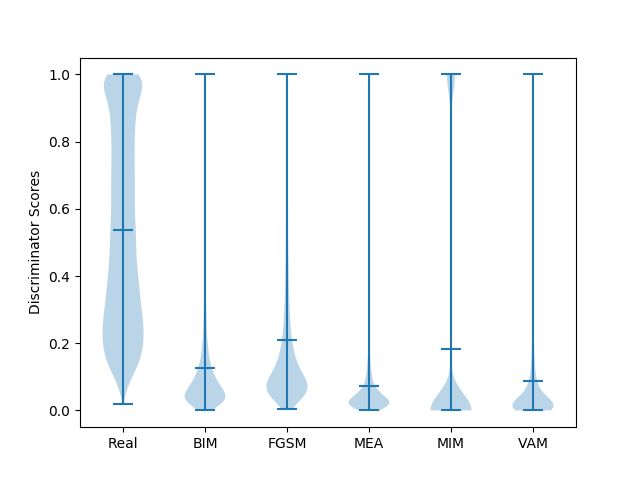} \\
(c) CIFAR-10 & (d) SVHN \\[7pt]
\end{tabular}
\caption{Distribution of the discriminator scores for different datasets. The violin denoted by real shows the discriminator scores on the validation set. The others show the scores of the discriminator after adding the perturbation to the images. As a result of the added perturbation, the images are pushed out of the data manifold and assigned a lower score. Note that not all perturbations lead to a successful adversarial sample, where successfulness is measured by the ability to cause miss-classification when using the target classifier.}
\label{fig:violins}
\end{figure*}

\paragraph{Results on Detection of Adversarial Samples.} 
We now directly test the hypothesis that the scores the discriminator assigns to samples are meaningful and can be utilized for the detection of adversarially perturbed images. Figure \ref{fig:violins} shows the discriminator distribution of these scores through violin plots for the different datasets and attack methods, respectively. Note that adversarial samples that do not cause miss-classification are included as well. The trend of consistently assigning higher scores to unmodified samples and lower scores to adversarial inputs is seen across all settings. Furthermore, we find that the score assigned by the discriminator to adversarial samples correlates to the severeness of the attack, i.e. attacks that cause larger decrease in the classification accuracy tend to be scored lower by the discriminator.\footnote{The severeness of the attack can be seen in Table \ref{tb:tablename} by comparing the accuracy of the classifier on the clean samples versus the accuracy of the classifier after applying the adversarial perturbations. } 
 
\begin{figure}[h]
\vskip 0.2in
\begin{center}
\centerline{\includegraphics[width=\columnwidth]{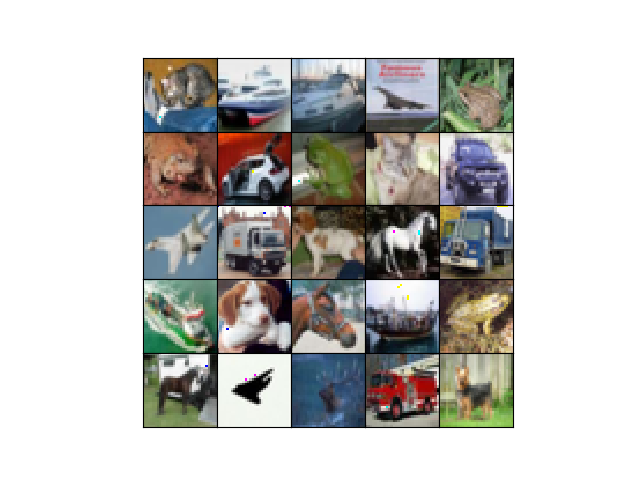}}
\caption{Randomly sampled images from the validation set of Cifar-10 for which the discriminator assigns low scores.}
\label{fig:cifarpatch}
\end{center}
\vskip -0.2in
\end{figure}

 For CIFAR-10 it is noticeable that some of the real images are assigned a low score by the discriminator. A randomly sampled batch of those is shown in Figure \ref{fig:cifarpatch}. As can be seen, the images tend to contain patches and unnatural artifacts.

Despite the fact that the GAN has only been trained on the real samples, and adversarial samples generated with a specific technique are never shown to the discriminator, it is capable of detecting adversarial perturbations irrespective of the attack method and dataset. This suggests that the discriminator behaves favorably in terms of ability to detect and distinguish adversarial samples.

\paragraph{Results on Cleaning Adversarial Samples.}
Once detected, the adversarial samples can in theory be cleaned by removing the perturbation such that the purified sample is assigned to the correct class when passed to the target classifier. Table \ref{tb:tablename} gives the classification accuracy on: (i) the entire test set, (ii) the adversarially perturbed samples and (iii) the cleaned samples. When used as a preprocessing procedure, the cleaning method significantly improves the classifier accuracy on the adversarial samples. The method generalizes well as reflected by the improved accuracy across the various dataset and attack method combinations.

\begin{figure*}[h]
\centering
\begin{tabular}{ccc}
  \includegraphics[width=55mm]{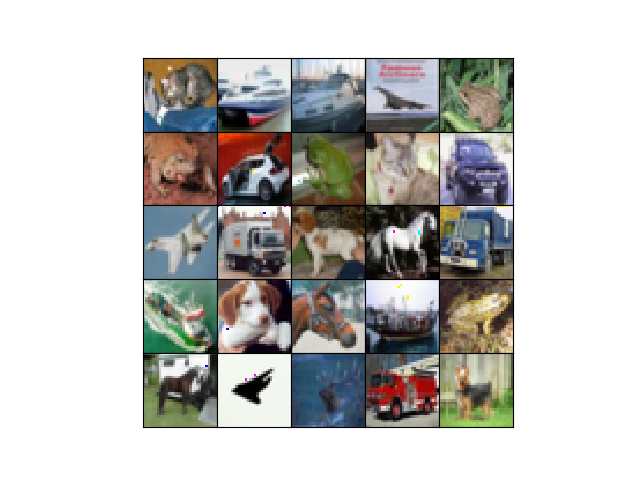} &   \includegraphics[width=55mm]{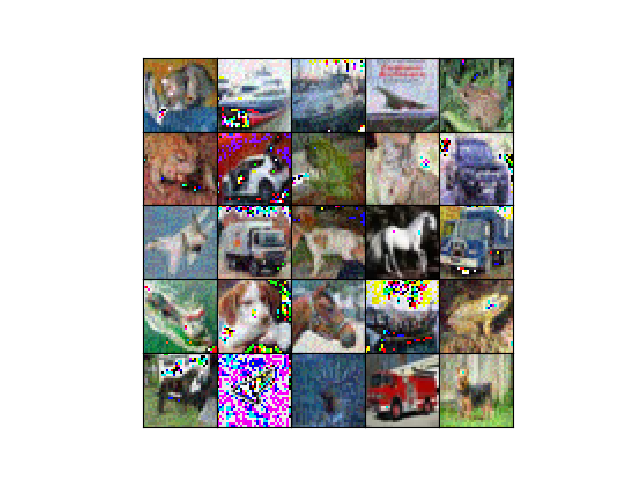} &   \includegraphics[width=55mm]{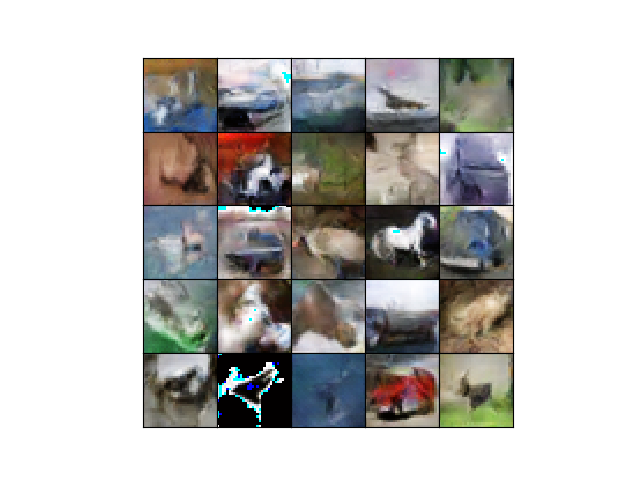}\\
(a) Real samples & (b) Adversarially perturbed samples & (c) Cleaned adversarial samples \\[6pt]
\end{tabular}
\caption{Here we show a few images from CIFAR-10 (a), then a few adversarial attacks performed on these images (b), and finally what they look like after cleaning by our proposed method (c).}
\label{fig:advcifar}
\end{figure*}

Figure \ref{fig:advcifar} showcases adversarially modified samples from CIFAR-10 and how they look like after our cleaning procedure has been applied.

In the following we investigate how specific aspects of the algorithm affect the results.

\subsection{Effect of the quality of the trained GAN}
Since we use different parts of a GAN to detect and clean adversarial samples, it is only natural that the quality of the GAN itself may influence the performance of these methods.

\begin{figure}[h]
\vskip 0.2in
\begin{center}
\centerline{\includegraphics[width=\columnwidth]{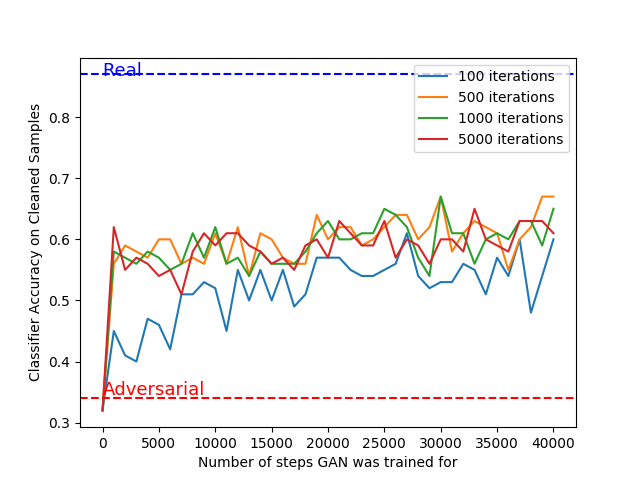}}
\caption{Effect of GAN quality on adversarial sample cleaning (FGSM ($l_\infty$) on SVHN): The classifier accuracy improves as the training of the GAN progresses. Different colors represent results for different number of cleaning steps.}
\label{fig:svhn-ganquality-clean}
\end{center}
\vskip -0.2in
\end{figure}

Figure \ref{fig:svhn-ganquality-clean} shows the classification accuracy where the discriminator and generator used for the cleaning procedure are trained for increasing number of training steps. As expected, we observe that as the GAN approaches convergence, the general trend of the performance quality increases. It can also be observed that the trends are not smooth, which can be attributed to the unstable nature of the GAN training.

\begin{figure}[h]
\vskip 0.2in
\begin{center}
\centerline{\includegraphics[width=\columnwidth]{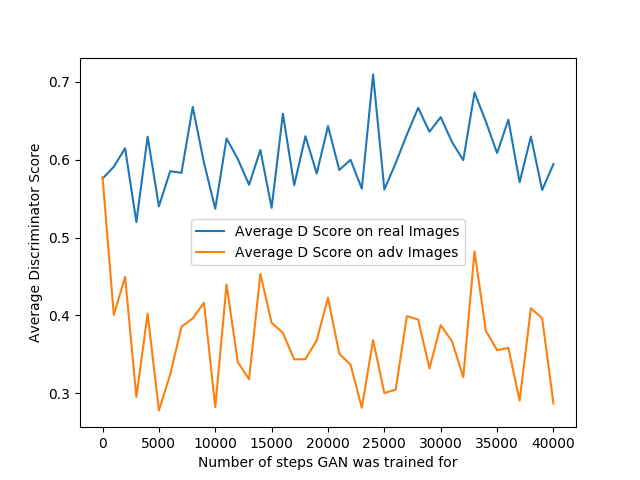}}
\caption{Effect of GAN quality on adversarial sample detection (FGSM ($l_\infty$) on SVHN): The average discriminator score for all the real and adversarial samples is given with the orange and blue curve, respectively. The discriminator is able to distinguish adversarial from real samples even early on throughout the training.}
\label{fig:svhn-ganquality}
\end{center}
\vskip -0.2in
\end{figure}

On the other hand, the performance of the detection mechanism of \emph{cowboy} is not influenced as much by the quality of the GAN, as shown in Figure \ref{fig:svhn-ganquality}. Even when the GAN is not fully trained, it assigns lower scores to adversarial in comparison to real samples, and hence, is able to detect the former.

\subsection{The Effect of The Two Terms}
\label{results}
In Section~\ref{sec:cowboy}, we argue that both the reconstruction term and the discriminator score of the cleaning objective are important for better purification of the adversarial samples. Table~\ref{tb:defensevcow} gives the comparison to Defense-GAN, a cleaning algorithm that is based only on the generator of the GAN. 

The results empirically justify the usefulness of the discriminator score in terms of pushing the adversarial samples on the data manifold.

\begin{center}
\begin{table*}[h]
\caption{Defense-GAN vs. Cowboy (5000 cleaning steps): Classifier accuracy on adversarial samples cleaned with each of the two methods. Cowboy consistently gives better results by the additional utilization of the discriminator score.}
{\small
\hfill{}
 \begin{tabular}{@{\extracolsep{1.5pt}}lrrrrrrrr}
 \toprule
 ATTACK & \multicolumn{2}{c}{ MNIST}  & \multicolumn{2}{c}{F-MNIST} & \multicolumn{2}{c}{CIFAR-10} & \multicolumn{2}{c}{SVHN} \\
\cline{2-3} \cline{4-5} \cline{6-7} \cline{8-9} 

& \multicolumn{1}{c}{Defense-GAN} & \multicolumn{1}{c}{Cowboy} & \multicolumn{1}{c}{Defense-GAN} & \multicolumn{1}{c}{Cowboy}  & \multicolumn{1}{c}{Defense-GAN} & \multicolumn{1}{c}{Cowboy} & \multicolumn{1}{c}{Defense-GAN} & \multicolumn{1}{c}{Cowboy} \\

 \midrule
FGSM ($l_{\infty})$  & 0.74 & 0.78      & 0.43 & 0.44       & 0.42 & 0.53       & 0.67 & 0.72 \\
PGDM ($l_{\infty})$  & 0.74 & 0.81      & 0.48 & 0.48       & 0.47 & 0.53       & 0.58 & 0.60 \\
BIM ($l_{\infty})$   & 0.12 & 0.15      & 0.11 & 0.13       & 0.38 & 0.47       & 0.49 & 0.49 \\
MIM ($l_{\infty})$   & 0.32 & 0.45      & 0.27 & 0.32       & 0.35 & 0.43       & 0.44 & 0.48 \\
VAM                  & 0.34 & 0.43      & 0.32 & 0.34       & 0.45 & 0.56       & 0.70 & 0.76 \\
\bottomrule
\end{tabular}
}
\hfill{}
\label{tb:defensevcow}
\end{table*}
\end{center}

\subsection{Summary}
For all the experiments, we use the standard GAN training procedure, without hyperparameter tuning and additional tricks for stabilizing the training (e.g. adding noise or regularizing \cite{roth2017stabilizing}, or using a more sophisticated GAN training and architecture, such as BEGAN \citep{berthelot2017began}). Even so, after training the GAN on the real samples, both the generator and discriminator contain useful information that can be leveraged for the defense against adversarial attacks.

The discriminator can be effectively used as a tool for detection of adversarial samples, whereas the combination of both the generator and discriminator constitutes a simple, yet powerful strategy for cleaning adversarial samples before their classification. As the method only works as a pre-processing step, it requires no modification of the training of the target classifiers, hence making it easily incorporable to already existing models.

\section{Conclusion}
In this paper, we presented \emph{cowboy}, a novel GAN-based method for successful detection and purification of adversarial samples. The method is based on the hypothesis that adversarial samples lie outside of the data manifold that is learned by a Generative Adversarial Network, irrespective of the underlying attack mechanism. We provide empirical evidence for this hypothesis and show that the discriminator acts as a good tool for the detection of adversarially perturbed samples. The defense strategy is based on projecting the detected adversarial samples back to the data manifold in order to clean the adversarial noise. Various experiments show the effectiveness and the generalization of the method across different attacks and datasets.

\FloatBarrier
\newpage
\newpage


\bibliographystyle{icml2018}
\bibliography{main}

\end{document}